\documentclass{article}

\usepackage{graphicx}
\usepackage{booktabs}
\usepackage{multirow}
\usepackage{arydshln}
\usepackage{tabularray}
\usepackage{booktabs}
\usepackage{longtable}
\usepackage{array}
\usepackage{listings}
\usepackage{float}
\usepackage{enumitem}   

\newcolumntype{L}[1]{>{\raggedright\arraybackslash}p{#1}}

\usepackage{siunitx}   

\usepackage[final]{neurips_2024}

\usepackage{caption} 

\usepackage{natbib}

\usepackage[utf8]{inputenc} 
\usepackage[T1]{fontenc}    
\usepackage{hyperref}       
\usepackage{url}            
\usepackage{booktabs}       
\usepackage{amsmath}        
\usepackage{amsfonts}       
\usepackage{nicefrac}       
\usepackage{microtype}      
\usepackage{xcolor}         

\usepackage{booktabs}      
\usepackage{pifont}        
\newcommand{\cmark}{\ding{51}}  
\newcommand{\xmark}{\ding{55}}  

\title{Arch-Router: Aligning LLM Routing with Human Preferences}

\author{%
  Co Tran \\
  Katanemo Labs, Inc. \\
  \texttt{co-tran@katanemo.com} \\
  \And
  Salman Paracha \\
  Katanemo Labs, Inc. \\
  \texttt{salman@katanemo.com} \\
  \And
  Adil Hafeez \\
  Katanemo Labs, Inc. \\
  \texttt{adil@katanemo.com} \\
  \And
  Shuguang Chen \\
  Katanemo Labs, Inc. \\
  \texttt{shuguang@katanemo.com} \\
}

\begin{document}

\maketitle
\begin{abstract}
With the rapid proliferation of large language models (LLMs) -- each optimized for different strengths, style, or latency/cost profile -- routing has become an essential technique to operationalize the use of different models. However, existing LLM routing approaches are limited in two key ways: they evaluate performance using benchmarks that often fail to capture human preferences driven by subjective evaluation criteria, and they typically select from a limited pool of models. In this work, we propose a preference-aligned routing framework that guides model selection by matching queries to user-defined domains (e.g., travel) or action types (e.g., image editing) -- offering a practical mechanism to encode preferences in routing decisions. Specifically, we introduce \textbf{Arch-Router}, a compact 1.5B model that learns to map queries to domain-action preferences for model routing decisions. Our approach also supports seamlessly adding new models for routing without requiring retraining or architectural modifications. Experiments on conversational datasets demonstrate that our approach achieves state-of-the-art (SOTA) results in matching queries with human preferences, outperforming top proprietary models. Our approach captures subjective evaluation criteria and makes routing decisions more transparent and flexible. Our model is available at: \texttt{https://huggingface.co/katanemo/Arch-Router-1.5B}.

\end{abstract}

\section{Introduction}
As new models continue to emerge rapidly \cite{llm_survey,naveed2023comprehensive}, users are shifting from single-model setups to multi-model systems to leverage the unique strengths of each LLM for tasks like text summarization, code generation, image editing \cite{hong2023metagpt, park2023generative}. LLM routing \cite{hybridllm,routerbench} has emerged as an effective way to build and deploy such systems by processing each user query through a router model that selects the most suitable LLM. 

However, existing routing approaches have limitations in real-world use. They typically optimize for benchmark performance while neglecting human preferences driven by subjective evaluation criteria \cite{stylepref, llm_subjective}. For instance, some routers are trained to achieve optimal performance on benchmarks like MMLU \cite{mmlu} or GPQA \cite{gpqa}, which don't reflect the subjective and task-specific judgments that users often make in practice. These approaches are also less flexible because they are typically trained on a limited pool of models \cite{mtbench}, and usually require retraining and architectural modifications to support new models or use cases.

This exposes a fundamental gap: the need for routing systems that align with subjective human preferences, offer more transparency, and remain easily adaptable as models and use cases evolve. To address this, we propose a preference-aligned routing framework—a principled approach to matching queries with routing policies based on user-defined preferences. In our framework, users define routing policies using a Domain-Action Taxonomy (e.g., healthcare, code explanation) expressed in natural language. Each policy is associated with a preferred model, enabling human-aligned control over routing decisions that capture subjective evaluation criteria grounded in real-world use. 

At the core of this framework is \textbf{Arch-Router} \footnote{Available at
\href{https://huggingface.co/katanemo/Arch-Router-1.5B}{https://huggingface.co/katanemo/Arch-Router-1.5B}.}, a compact 1.5B language model that matches user queries to routing policies with high accuracy. We also introduce a complementary data creation pipeline that produces high-quality, labeled conversations—capturing nuanced intents and complex dialogue patterns—to train and evaluate preference-aligned routing effectively.

Given a set of natural language policy descriptions, Arch-Router makes accurate routing decisions without requiring retraining or architectural changes—making our framework highly adaptable as new routes or models are added. Trained on rich conversational data, Arch-Router handles diverse conversations and multi-turn interactions more effectively than static embedding-based methods \cite{routerdc, jang2023exploring}. Furthermore, our experiments show that Arch-Router outperforms top proprietary LLMs by 7.71\% on average.

To summarize, our work makes the following key contributions:
\begin{enumerate}
    \item A preference-aligned routing framework comprising a Domain-Action Taxonomy and 1.5B model that maps queries to user-defined routing policies with high accuracy.

    \item Our approach aligns with subjective human preferences, enabling more practical and user-relevant routing decisions.

    \item It offers transparency and flexibility in model routing, reflecting how LLMs are evaluated, integrated, and applied in real-world scenarios.

\end{enumerate}
\section{Related Work}
Recent LLM routing work is focused on two broad categories: task-specific routing or performance-based routing. We provide an overview of these methods and highlight the gaps that motivate our work in guiding routing decisions with human preferences.

\paragraph{Task-Based Routing.}
This approach focuses on directing queries to models or systems specialized for a predefined task. These strategies range from using BERT-based classifiers for domain classification \cite{modem} to more sophisticated methods involving k-NN layers and entropy-based classification \cite{jain2024composition}. For example, OrchestraLLM \cite{orchestrallm} implements a retrieval-based router: at inference it finds the kk most similar dialogue exemplars via embedding similarity and routes to a small or large expert model by majority vote. HuggingGPT uses LLMs for model selection based on model descriptions \cite{hugginggpt}. This principle has also been extended to Retrieval-Augmented Generation (RAG) scenarios, where routers dynamically select the appropriate data modality for retrieval to improve response accuracy \cite{benchmarking_multimodal,route_rag,universalrag}. While effective for well-defined and clearly separable tasks, these approaches struggle in settings where task boundaries blur or overlap. They are particularly brittle in multi-turn interactions, where the user’s intent may evolve or drift, requiring adaptive reclassification that most static routers are not designed to handle without retraining or manual intervention.

\paragraph{Performance-Based Routing.}
The most prevalent research in LLM routing is around performance-based approaches that optimize for cost-performance trade-offs in routing between different LLMs. This approach typically uses a scoring function to predict which model will yield the best possible performance for a given query. They aim to automatically select the model most likely to produce a high-quality response, often by training a router to decide whether a query can be handled by a "weak," cheaper model or must be escalated to a "strong," more expensive ones. Examples are RouteLLM\cite{ong2024routellm} and HybridLLM\cite{hybridllm}. Or FrugalLLM that expands further by choosing from a larger pool of LLMs with specific budget constraints  \cite{frugalgpt}. Training methods range from leveraging historical performance statistics \cite{feng2024graphrouter, xia2024llm} to semantic similarity \cite{routerdc, jang2023exploring} and ELO-based ranking \cite{zhao2024eagle}.

While achieving the best possible performance is the right end goal, current work suffers from limitations that hinder its use in practical real-world settings. First, these routers are often brittle and rigid. Trained on a small, limited set of models (typically 3-10) and specific task domains (e.g., coding, math) using benchmarks like RouterBench~\cite{routerbench} or MT-Bench~\cite{mtbench}. Their performance degrades on out-of-domain queries and they cannot adapt to new models without retraining~\cite{llm_robust}. Second and more fundamentally, current work treats quality as an objective measurement, neglecting human preferences driven by subjective evaluation criteria. As a consequence, performance-based routing is more suitable in controlled environments with stable model sets and objective tasks, but ill-suited for real world scenarios where subjective evaluation of response quality matter.

\paragraph{Human Preferences.} Preference modeling and Reinforcement Learning from Human Feedback (RLHF) have led to notable gains in LLM performance on well-defined tasks like question answering and summarization \cite{rlhf, ppo, steerpreference, worldpm}. However, these methods are less effective in scenarios where quality is shaped by subjective factors—such as tone, style, or task-specific expectations—that cannot be reliably captured by well-defined benchmarks \cite{stylepref}. Studies consistently show a disconnect between LLM-based evaluators and human judgments in such settings \cite{llm_subjective, llm_judge_limitation, ye2024justice}. While LLMs can mimic human-like fluency, they often overlook subtle signals in user intent and encode their own systemic biases \cite{ye2024justice}. These limitations undermine routing methods that rely on automatic quality scores for model selection. Such systems are typically trained on LLM-labeled preference datasets and evaluated on benchmarks that reflect aggregate rather than individual preferences, such as Chatbot Arena \cite{mtbench} and RouterBench \cite{routerbench}. In contrast, our work reframes LLM routing as a preference-alignment problem, where model selection is guided by matching queries to user-defined policies expressed via a Domain (e.g., finance) or Action taxonomy (e.g., image search). This offers a practical mechanism to encode subjective preferences directly into routing decisions, making them more interpretable, flexible, and aligned with real- world usage.

\section{Problem Formulation}
\label{problem_formulation}

\subsection{Preliminary}
The main goal of LLM routing is to choose the appropriate model for a given query from a pool of models, creating a system that is more capable and effective than any single model. LLM Routing is commonly defined as a function $\mathcal{R}: (q, \mathcal{P}) \rightarrow M$ that maps a user query $q \in \mathcal{Q}$ to an appropriate model $M$ from a model pool $\mathcal{M}$ under routing objectives $\mathcal{P}$ such as optimal performance and low cost.

\subsection{Preference-aligned Routing}
To incorporate human preferences as routing objectives, we introduce a routing framework that decouples route selection from model assignment. We define a set of \textit{route policies} $\mathcal{C} = \{c_1, \ldots, c_k\}$, where each route policy $c_i = (n_i, d_i)$ is a tuple consisting of a unique route identifier $n_i$ paired with a natural language description $d_i$. Additionally, we define a mapping $\mathcal{T}: \mathcal{C} \rightarrow \mathcal{M}$ that associates each route policy with a specific model. 

\paragraph{Domain–Action Taxonomy} \label{taxonomy}
In this work, we focus on modeling LLM routing to align with human preferences driven by subjective evaluation criteria. To incorporate human preferences as routing objectives, we adopt a Domain–Action taxonomy, a two-level hierarchy that mirrors how people typically describe tasks—starting with a general topic and narrowing to a specific action. \textit{\textbf{Domain}} (e.g., legal and finance) captures the high-level thematic context of a query while \textit{\textbf{Action}} (e.g., summarization and code generation) denotes the specific operation requested. This taxonomy serves as a mental model to help users define clear and structured routing policies. Separating Domain and Action strikes a balance between expressiveness and simplicity: it avoids an unwieldy flat list of composite labels (e.g., finance summarization and legal advice) and introduces a natural fallback. If a query is too vague to match an Action, the router can still resolve the Domain—maintaining robustness and reducing semantic ambiguity.

\begin{figure}[!t]
  \centering
  \includegraphics[width=14cm]{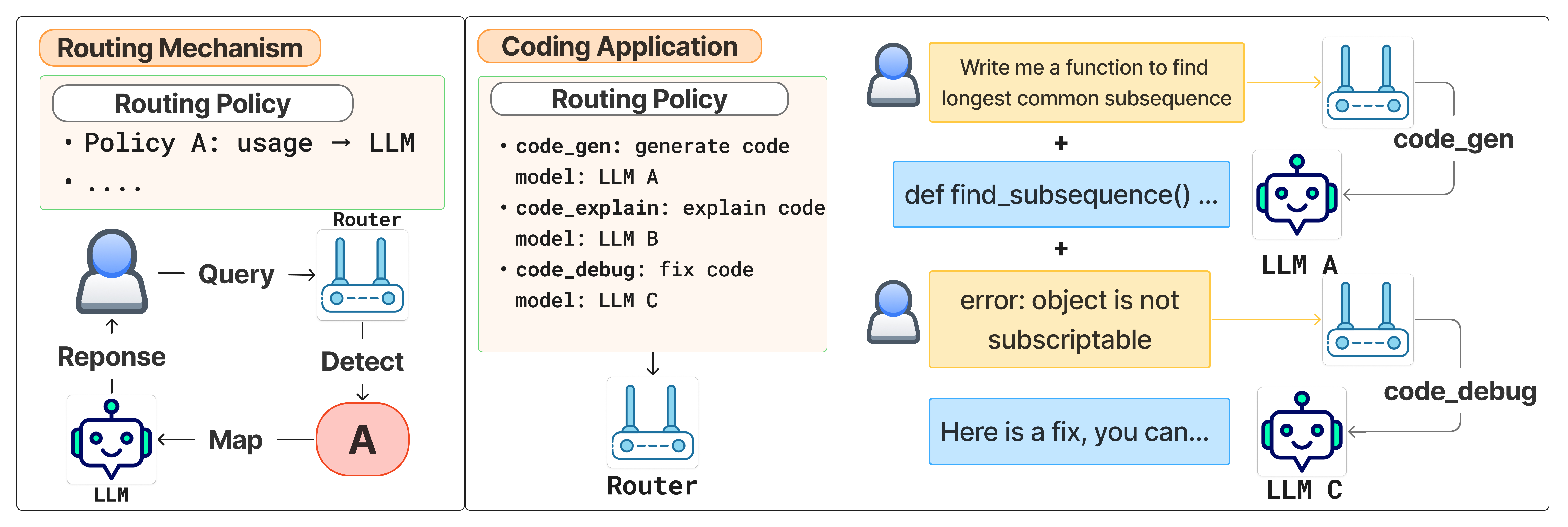}
  \caption{Preference-Aligned Routing Mechanism. The routes policies and user conversation is provided to the router to select the appropriate policy and corresponding LLM. Example of usage in a coding application is shown in the right.}
  \label{routing_workflow}
\end{figure}

\paragraph{Routing Mechanism}
As shown in Fig. \ref{routing_workflow}, the routing process includes two stages: first, a preference-aligned router $ \mathcal{F}: (q,\mathcal{C}) \rightarrow c$ takes a user query $q$ and the complete set of route policies $\mathcal{C}$, then selects the most appropriate $c \in \mathcal{C}$ based on the policy descriptions. Second, $\mathcal{T}$ maps the selected route policy to its associated LLM model: $M = \mathcal{T}(c)$ \footnote{An open-source implementation is available at \href{https://github.com/katanemo/archgw}{github.com/katanemo/archgw}.} encodes human preferences in both the construction and description of route policies in $\mathcal{C}$, and the associations between route policies and models defined in $\mathcal{T}$. Thus, 'best' is defined directly by human preference rather than by an estimated performance score. Because the LLM routing process is decoupled to $\mathcal{R} = \mathcal{T} \circ \mathcal{F}$, model selection is completely delegated to \(\mathcal{T}\).
Models can therefore be added, removed, or swapped by editing \(\mathcal{T}\) alone, without retraining or modifying the router. This decoupling provides the flexibility required in practical deployments where model usage evolves continuously.

\paragraph{Practical Benefits}
This preference-aligned routing design offers several key advantages over traditional routing approaches. First, the decoupling of route selection from model assignment enables dynamic reconfiguration without retraining the router—users can update model mapping in $\mathcal{T}$ to accommodate new models, changed requirements, or performance optimizations while preserving the learned routing logic in $\mathcal{F}$. Second, the natural language descriptions $d_i$ in route policies make the system interpretable and auditable, allowing users to understand routing decisions and validate that policies align with intended preferences. Third, the Domain-Action taxonomy provides both flexibility and structure: it supports fine-grained control when specific domain-action combinations are defined, while gracefully degrading to domain-level routing when actions are ambiguous or undefined. Finally, this framework naturally accommodates human-in-the-loop refinement, as route policies can be iteratively adjusted without requiring architectural changes or model retraining, making the system both maintainable and adaptable to evolving preferences.

\section{Methodology}
\label{method}

We introduce Arch-Router, a compact 1.5B generative language model designed for our framework. 
Given a query \(q\) and a complete policy set \(\mathcal{C}\), Arch-Router generates the identifier of the best-matching policy.  Because the full set of policy descriptions is included in the prompt, the model can adopt new or modified routes at inference time without retraining, aligning seamlessly with the framework’s modular design. Furthermore, we present a novel, two-phase data creation pipeline  to model the routing framework. Our data pipeline (i) builds a rich corpus of route policies and realistic multi-turn dialogues and (ii) injects real-world scenarios—topic shifts, irrelevant turns, and noise.

\subsection{Data Creation Framework}

\begin{figure}[!t]
  \centering
  \includegraphics[width=14cm]{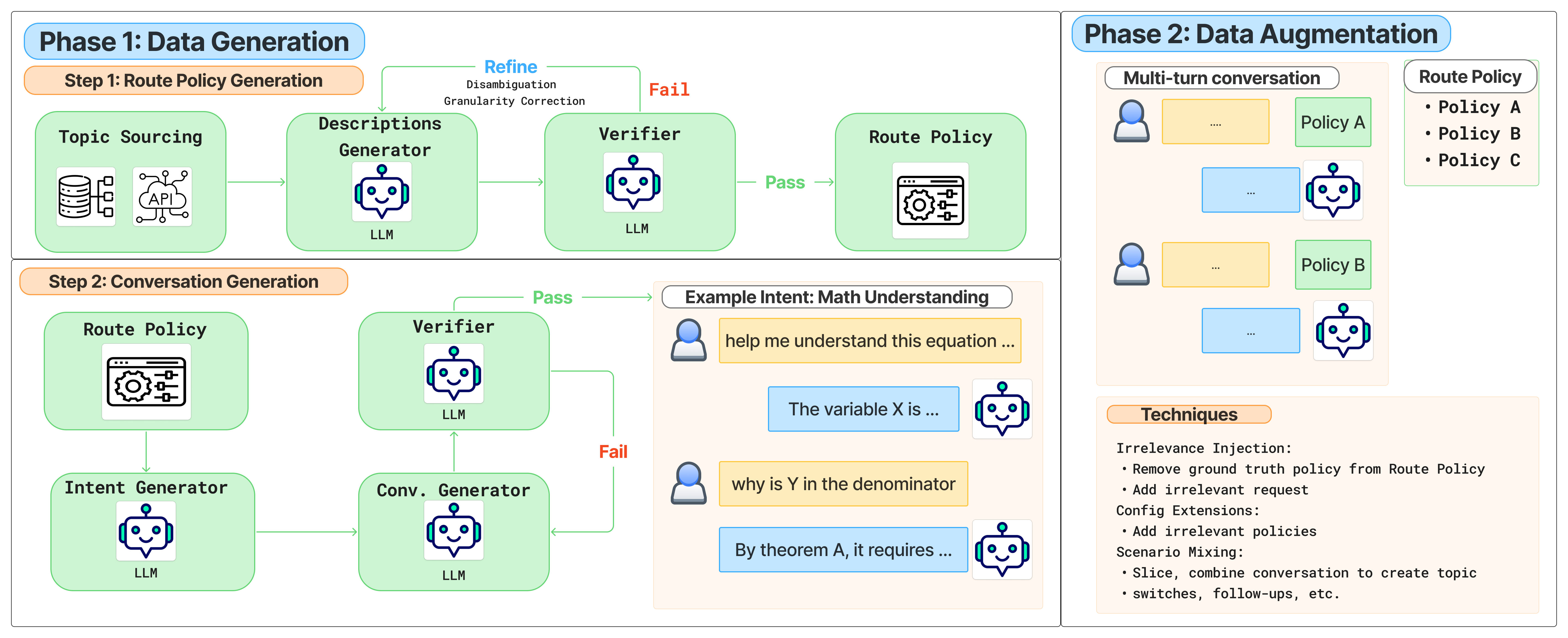}
  
  \caption{Overview of the data creation for Arch-Router framework. Phase 1 generates route configurations through an LLM process with feedback loops. Phase 2 generates conversations from generated intent. Phase 3 augments the conversations to get diverse scenarios and irrelevance.}
  \label{fig:data_framework}
\end{figure}

To support Arch-Router, we generate a corpus that represents the operating environment of the routing function in two phases (see Fig. \ref{fig:data_framework} for full illustration). Phase 1 focuses on generating clean, structurally correct conversations grounded in a verified set of route policies. Phase 2 then systematically introduces real-world complexities to augment these conversations and policy set, reproducing the ambiguity the router must resolve in practice. This two-phase separation is crucial: it allows us to first ensure the correctness and quality of the core conversational data, and then independently layer on challenges to improve data robustness. Every data sample contains (i) the conversation, (ii) the full set \(\mathcal{C}\) as route policies, and (iii) the ground truth policy, yielding a data set that represents the preference-aligned routing task as it occurs during inference time.

\paragraph{Data Generation.}
Phase 1 produces clean, labeled conversations through a structured two-step process. First, we generate route policies by constructing a diverse topic pool from industry classifications \cite{morningstar_equity_methodology}, academic benchmarks such as \textsc{MMLU} \cite{mmlu}, and real-world API documentation \cite{apigen}. An LLM is used to generate candidate route policies from this pool. These candidates are then validated and refined by a second LLM to ensure clarity, appropriate granularity (as defined by the Domain-Action Taxonomy \ref{taxonomy}), and semantic coherence. Second, we synthesize conversations using the curated set of route policies. For each policy, an LLM generates a specific conversational intent, which is then passed to another LLM to produce a full dialogue. To ensure data quality, a final LLM verifies the alignment between the conversation and the intended routing policy. Conversations that fail this check are regenerated.

\paragraph{Data Augmentation.}
Phase 2 enhances the dataset by systematically incorporating real-world complexity to improve the robustness of the routing model. We apply three augmentation techniques to diversify conversational patterns: 1) Irrelevance injection introduces noise by adding off-topic user messages or removing the ground-truth route policy from a sample, simulating ambiguity in user intent, 2) Policy modification alters the candidate set of route policies by including irrelevant or misleading options, creating more challenging decision boundaries, and 3) Scenario mixing enriches the data by combining segments from different conversations to create longer dialogues with abrupt topic shifts, follow-up questions, and abandoned intents. These augmentations yield a more representative and challenging dataset, essential for training a routing model that generalizes well to real-world usage.


\subsection{Arch-Router}
We construct Arch-Router - $\mathcal{F_{\text{Arch}}}$ as a generative language model where it is trained to generate the route identifier of the target route policy. $\mathcal{F_{\text{Arch}}}$ is provided with a structured prompt $x$ (for full prompt template see \ref{tab:system-prompts}) that contains both the user query $q$ and the set of all possible route policies $\mathcal{C}$. 

The model $\mathcal{F_{\text{Arch}}}$ processes this input prompt $x$ to generate a route identifier. The training objective is maximizing the generated output being the correct route policy, $c_{\text{true}}$, which represented as minimizing the cross-entropy loss over $(x, c_{\text{true}})$ pair \cite{sft}:
\begin{equation}
 \arg\min_{\mathcal{F_{\text{Arch}}}}  \mathbf{\mathcal{L}}(\mathcal{F_{\text{Arch}}}(x), c_{\text{true}})
\end{equation}

\paragraph{Design Benefits.} The decision to implement Arch-Router as a generative language model offers advantages over other approaches like classifier or semantic matching. Unlike a multi-class classifier, which is architecturally bound to a fixed set of output classes, generative model is inherently flexible. By providing the available route policies as part of the input prompt, Arch-Router can adapt to new policies at inference time without retraining. Moreover, this approach allows Arch-Router to leverage its foundation model’s pre-trained knowledge, improving semantic understanding of both the query and the policy descriptions. Furthermore, this approach allows the model to process the entire conversation history together with the policy descriptions simultaneously. This is a distinct advantage over semantic search methods or  classifier that embed the query and routes separately.


\section{Experiment}
Based on the data creation framework, we curated a set of 43k samples to train the Arch-Router model. In this section, we lay out the details and results of our training and evaluation, consequently demonstrating the robust design of data creation framework and the effectiveness of preference-aligned routing.
\subsection{Experiment Setup}

\paragraph{Evaluation Corpora.}
We evaluate our model on four public datasets described in below. For all the dataset, we randomly sample a portion from each dataset for eval, the full details are shown in Table \ref{table:dataset_statistic}. To adapt the evaluation sets into our preference-aligned routing framework, we augmented each dataset. Additionally, all route descriptions are generated using an LLM and went through verification step for consistency with the training dataset.
\begin{itemize}
\item \textbf{CLINC-150} \cite{clinc150} convers a broad range of user needs in task-oriented dialogue systems. Since the dataset doesn't have the hierarchy between action and domain, we leverage LLM to cluster them by intent name semantic.

\item \textbf{MANtIS} \cite{mantis} is a multi-domain, human-human dialogue dataset focused on information-seeking tasks. We only use the domains provided from this dataset to measure performance.

\item \textbf{SGD} \cite{sgd} (Schema-Guided Dialogue) is a dialogue state tracking dataset. Similarly to CLINC-150, we cluster action level route policies by LLM to create domains ground truth.

\item \textbf{LMSYS-1M} \cite{lmsys} is a large-scale evaluation dataset with over one million conversations. For our evaluation, we annotated each conversation with route policy labels and configurations using both LLM and human. 

\end{itemize}
\paragraph{Comparison Pool.} We compare the performance of our trained model Arch-Router against state-of-the-art
proprietary model familes such as OpenAI (GPT-4o-mini, GPT-4o); Anthropic (Claude-3.5-haiku,
Claude-3.7-sonnet); Gemini (Gemini-2.0-flash, Gemini-2.0-flash-lite), and the original Qwen2.5-1.5B.
\paragraph{Base Model.} 
Considering the balance between speed and performance for practical deployment, we use models under $2B$ in size from the following families: Qwen 2.5 \cite{qwen2}, Qwen 3 \cite{qwen3}, Llama 3.2 \cite{llama}, Gemma 2\cite{gemma2}, and Gemma 3\cite{team2025gemma}. The best performing model based on validation split is chosen to be Arch-Router, which based on Qwen 2.5 - 1.5B.
\paragraph{Training Details.} 
We adopt Supervised fine-tuning (SFT) \cite{sft}, carried out with the Llama-Factory library \cite{llama_factory}, using full-parameter updates in bfloat16 precision, the AdamW optimizer, and a maximum of four epochs on a single NVIDIA A100 GPU. We train over $43,000$ pairs of samples with training/validation split as 90/10. Full training data details are shown in Table \ref{table:dataset_statistic}.

\paragraph{Metrics.} 
We measure model performance at the following three levels of accuracy
\begin{enumerate}
    \item \textbf{Turn}: the fraction of single turns for which the predicted route matches the ground-truth.
    \item \textbf{Span}: the fraction of contiguous spans of identical route labels in which \emph{all} turns are predicted correctly; labels immediately outside the span may differ.
    \item \textbf{Conversation}: the fraction of conversations whose every turn is correctly predicted.
\end{enumerate}
For a more comprehensive evaluation, we also include these three scenarios:

\begin{enumerate}
    \item \textbf{Fine-grained Query \(\rightarrow\) Fine-grained Answer (fQfA).} the model's ability to match the exact action route policy. 
    \item \textbf{Fine-grained Query \(\rightarrow\) Coarse-grained Answer (fQcA).} the model's ability to match the domain route policy where there is no exact action route policy present.
    \item \textbf{Irrelevance.} the model's ability detect irrelevant request or request is complete.
\end{enumerate}

\begin{table}[!t]
\centering
\begin{tabular}{lcccc}
\toprule
\textbf{Models} & \textbf{Turn} & \textbf{Span} & \textbf{Conv.} & \textbf{Overall} \\
\midrule

Qwen2.5-1.5B       & 34.37 & 16.09 & 11.61 & 20.69 \\
\midrule
GPT-4o                          & 93.96 & 90.74 & 84.52 & 89.74 \\
GPT-4o-mini                     & 87.49 & 80.89 & 80.00 & 82.79 \\
Claude-sonnet-3.7            & \textbf{96.24} & 94.70 & 87.45 & 92.79 \\
Claude-haiku-3.5             & 88.21 & 81.46 & 83.72 & 84.96 \\
Gemini-2.0-flash            & 89.19 & 85.05 & 85.79 & 85.63 \\
Gemini-2.0-flash-lite       & 85.13 & 72.61 & 74.76 & 76.69 \\
\midrule

\textbf{Arch-Router}              & 96.05 & \textbf{94.98} & \textbf{88.48} & \textbf{93.17} \\

\bottomrule
\\
\end{tabular}
\caption{Overall routing performance across different evaluation granularities. Each metric is an aggregated score averaged over all test datasets and scenarios (fQfA, fQcA, Irrelevance) from Tables \ref{clinc_table},\ref{mantis_label} ,\ref{sgd_table} ,\ref{lmsys_table} in Appendix \ref{appendix_a}. The table shows how accuracy is maintained as the evaluation scope expands from a single turn, to a multi-turn span, to the entire conversation.}
\label{tab:overall_table}
\end{table}

\subsection{Results}
\label{results}
\begin{table}[!t]
    \centering
    
    \begin{tabular}{lcccc}
        \toprule
        \textbf{Models} & \textbf{fQfA} & \textbf{fQcA} & \textbf{Irrelevance} & \textbf{Turn Overall} \\
        \midrule
                Qwen2.5-1.5B               & 51.74          & 40.89          & 10.48    & 34.37      \\
        \midrule
        GPT-4o                 & 97.09          & 91.85          & 92.94      &93.96    \\
        GPT-4o-mini            & 95.77          & 86.03          & 80.67       &87.49   \\
        Claude-sonnet-3.7      & 96.87          & \textbf{94.63} & \textbf{97.21} & \textbf{96.24}\\
        Claude-haiku-3.5       & 93.85          & 90.27          & 80.51        & 88.21  \\
        Gemini-2.0-flash   & 93.74          & 88.72          & 85.11      & 89.19    \\
        Gemini-2.0-flash-lite  & 92.93          & 87.61          & 74.86     &85.13     \\
        \midrule

        \textbf{Arch-Router}             & \textbf{98.11} & 93.56          & 96.49     & 96.05     \\
        \bottomrule
        \\
    \end{tabular}
    \caption{Detailed breakdown of turn-level performance across three evaluation scenarios which also aggregated from Tables \ref{clinc_table},\ref{mantis_label} ,\ref{sgd_table} ,\ref{lmsys_table} in \ref{appendix_a}. The scores show model accuracy on the action-level task (fQfA), on domain-level when action is not available (fQcA), and irrelevant queries (Irrelevance). The Turn Overall column corresponds to the 'Turn' column in Table \ref{tab:overall_table}.}
    \label{tab:turn_level_accuracy_by_scenario}
\end{table}

Arch-Router records the highest overall routing score of \(93.17\%\) (Table \ref{tab:overall_table}), surpassing every other candidate model on average by $7.71\%$.   Its margin widens with context length: per-turn accuracy is competitive, yet span-level and full-conversation accuracies rise to the top \(94.98\%\) and \(88.48\%\), respectively—evidence that the model can follow multi-turn context better than other candidate models.
A scenario-wise breakdown (Table \ref{tab:turn_level_accuracy_by_scenario}) sharpens this picture.  On the fine-grained fQfA benchmark Arch-Router attains \(98.11\), confirming its ability to map queries to specific action routes.  
Performance on the other tasks— fQcA and irrelevance—matches Claude-sonnet-3.7, the strongest proprietary model in our candidate set, indicating that the generative, preference-aligned design remains robust even when the query cues are coarse or out-of-distribution.

\section{Analysis}
While the aggregated performance scores in the previous section demonstrate the effectiveness of our approach, this section uncovers the qualitative and practical implications of our framework. We begin by examining the specific error patterns to understand in which scenarios Arch-Router succeeds and fails. Then, we contextualize our work by comparing the preference-aligned to performance-based routing and discussing the inherent limitations of our approach.

\subsection{Analysis on Model Behavior}
\begin{figure}[t]
  \centering
  \begin{minipage}[t]{0.48\linewidth}
    \centering
    \includegraphics[width=\linewidth]{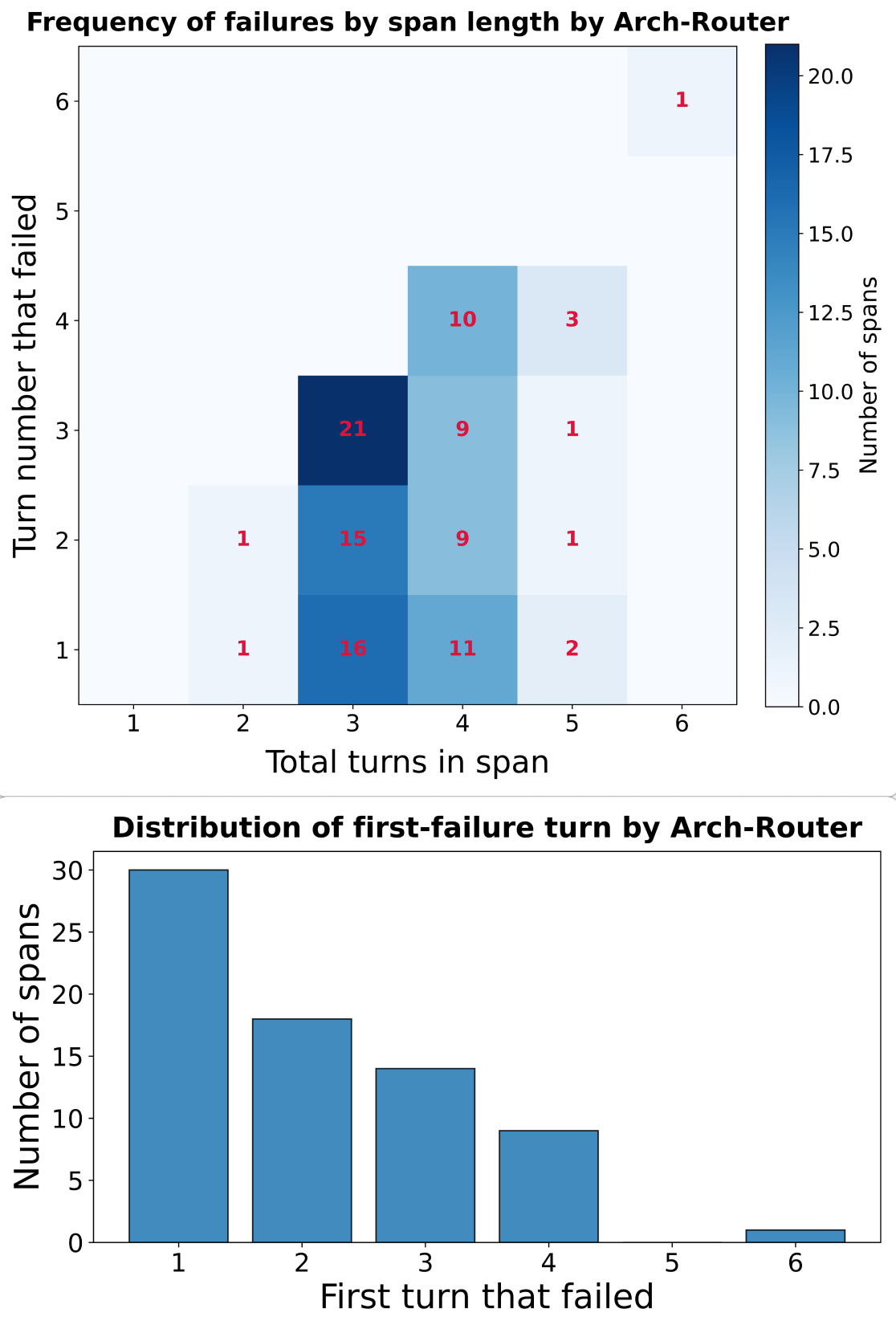}
    \label{fig:first}
  \end{minipage}
  \hfill                 
  \begin{minipage}[t]{0.49\linewidth}
    \centering
    \includegraphics[width=\linewidth]{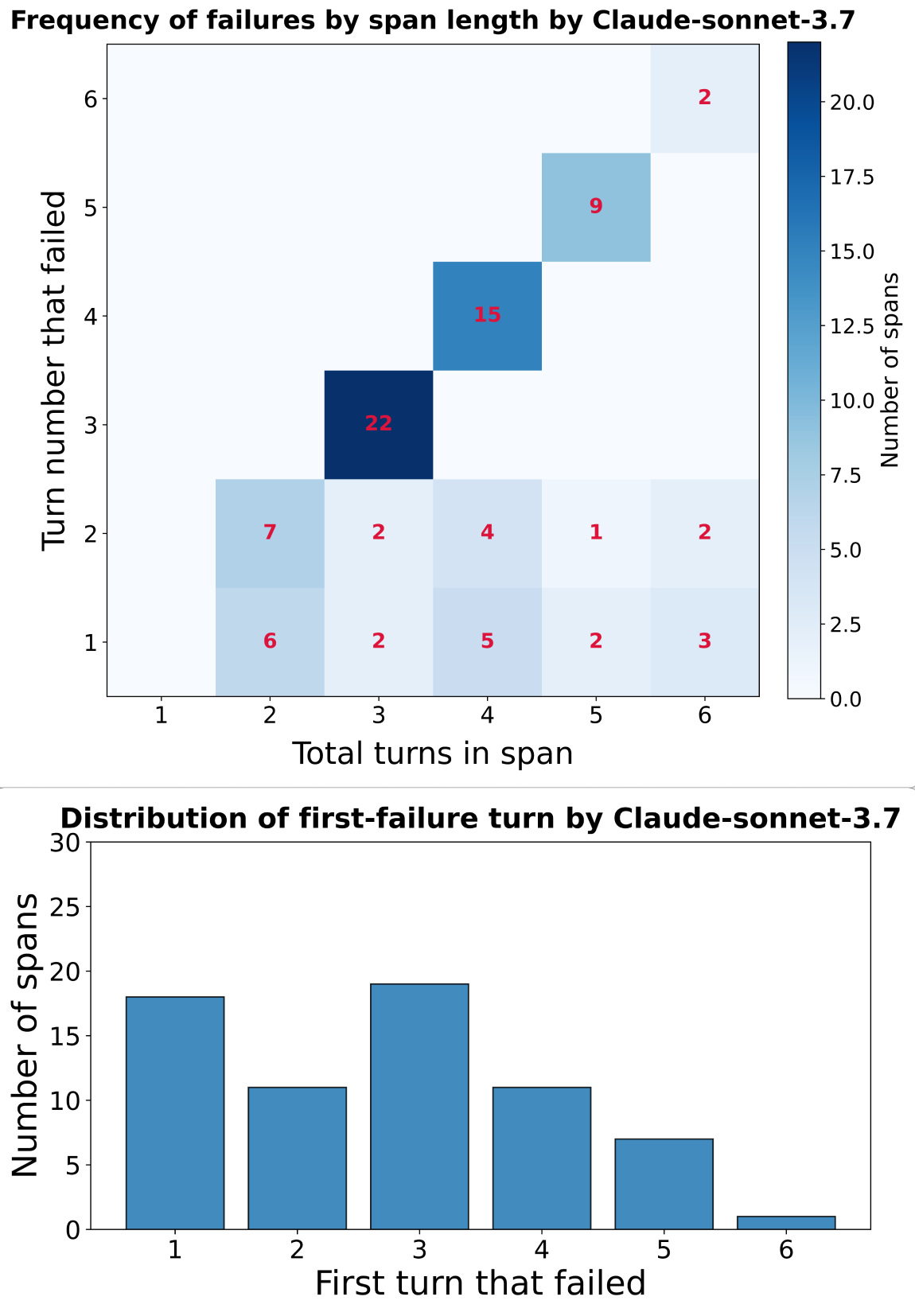}

    \label{fig:second}
  \end{minipage}
    \caption{Comparison of failure distributions for Arch-Router (left) and Claude-Sonnet-3.7 (right) on SGD dataset \ref{sgd_table}. Each panel combines a heat-map and a histogram. The heatmaps illustrate at which turn multi-turn conversational spans tend to fail and how often different span lengths are affected. The histograms show the count distribution of the first fail turn index.}  \label{fig:error_pattern}
\end{figure}

In Fig. \ref{fig:error_pattern}, we provide an analysis on error patterns for both Arch-Router and Claude-Sonnet-3.7 
A closer look at the error patterns reveals distinct failure profiles. In general, both models exhibit the highest frequency of failures in spans of moderate length (3-4 turns), as shown in the heatmaps. However, the distribution of when these failures first occur highlights a crucial behavioral difference. Arch-Router fails most often on the first turn. The bar chart shows a large spike in first-failures at turn 1, which then steadily decreases. This suggests that Arch-Router is most vulnerable to initial query ambiguity; if it correctly identifies the user's intent from the start, it is highly robust and unlikely to fail on subsequent follow-up turns.

In contrast, Claude-Sonnet-3.7's failures are more evenly distributed throughout the span. Its bar chart shows that the highest number of first-failures occurs mid-span at turn 3, with significant failures also happening at turns 1. This pattern suggests that Claude-Sonnet-3.7 is less sensitive to ambiguity in the first turn but more likely to fail on later follow-up requests. After correctly handling the first turn, it can drift off-context or misread a nuanced follow-up, leading to errors later in the span. The error pattern analysis further reinforce our observation on the strong ability of Arch-Router on tracking continuous intents.   

\subsection{Preference-Aligned V.S. Performance-Based Routing}

Preference-aligned and performance-based routing represent two fundamentally different approaches for routing LLMs, distinguished by their routing objectives and mechanisms. Preference-aligned routing frames as multi-class classification over user-defined route policies. In contrast, performance-based routing treats it as constrained optimization, seeking to select a model that maximizes a predicted numerical quality score while adhering to a budget, as formulated in \cite{jitkrittum2025universal}.

This reflects a fundamental difference in where the judgment of quality resides. In preference-aligned routing, the user makes the quality judgment, defining a set of route policies that serve as a categorical proxy. The router's task is not to guess what LLM is best for each task, but to faithfully map a user query to the user-defined policies. Conversely, performance-based routing delegates the quality judgment to the model, using a score function as a direct numerical proxy to predict the best outcome. This leads to a natural difference in usage: Consequently, preference-aligned routing is the method of choice when the quality is subjective and driven by human—that is, when success is determined by user-defined criteria rather than by an automatic metric—whereas performance-based routing best suits cost-sensitive settings equipped with reliable, model-predicted quality scores.

\subsection{Limitations}

Preference-based routing offers clear advantages in transparency and control, but this design introduces two limitations. First, routing accuracy is constrained by the precision of the policy set; ambiguous or overlapping policy descriptions directly degrade routing performance. For instance, if one policy is defined as ``review contract clauses'' and another as ``analyze legal documents,'' a query about non-disclosure agreements could reasonably match both descriptions, leading the router to make arbitrary routing decisions that may not align with user intent. Second, overall routing effectiveness is inherently constrained by user model selection. If users assign inappropriate models to their route policies, even accurate routing will result in suboptimal outcomes.


\section{Conclusion}
In this work, we introduce a preference-aligned routing framework that lets users encode human preferences as explicit route policies and decouple those policies from model selection -- enabling routing decisions that reflect subjective evaluation criteria while improving transparency and flexibility. Within this framework, we presented Arch-Router, a compact 1.5B generative language model and a two-phase data-creation pipeline. Empirical results on multi-turn conversations benchmarks show that Arch-Router surpasses the best proprietary models. From our results and analysis, we highlight the effectiveness and practical use of human preferences as a primary routing objectives. Future work includes exploring these following directions: 1) hybrid frameworks that combine preference-aligned routing with precise performance objectives, and 2) human preference modeling to support a wider range of routing policies.
Addressing these challenges will broaden the applicability of preference-aligned routing and reinforce its role as a practical approach for LLM routing.

\bibliographystyle{plain}  

\bibliography{router_bib}
\newpage
\appendix
\section{Datasets Performance}
\label{appendix_a}
\begin{table}[ht]
\centering\small
\begin{tabular}{@{}p{3.5cm}p{10cm}@{}}
\toprule
\textbf{Prompt element} & \textbf{Exact text} \\
\midrule
TASK &
\begin{minipage}[t]{\linewidth}\ttfamily
You are a helpful assistant designed to find the best suited route.\newline
You are provided with route description within \textless routes\textgreater\textless/routes\textgreater\ XML tags:\newline
\textless routes\textgreater\newline
\textbackslash n\{routes\}\textbackslash n\newline
\textless/routes\textgreater\newline\newline
\textless conversation\textgreater\newline
\textbackslash n\{conversation\}\textbackslash n\newline
\textless/conversation\textgreater
Your task is to decide which route is best suit with user intent on the conversation in \textless conversation\textgreater\textless/conversation\textgreater\ XML tags.\newline
Follow the instruction:\newline
1.~If the latest intent from user is irrelevant or user intent is full filled, respond with other route \{"route": "other"\}.\newline
2.~Analyze the route descriptions and find the best match route for user latest intent.\newline
3.~Respond only with the route name that best matches the user's request, using the exact name in the \textless routes\textgreater\ block.\newline\newline
Based on your analysis, provide your response in the following JSON format if you decide to match any route:\newline
\{"route": "route\_name"\}
\end{minipage}\\[2ex]
\\
\bottomrule
\\
\end{tabular}
\caption{Routing Prompt}
\label{tab:system-prompts}
\end{table}

This section provides the detailed performance breakdown for each model across our four evaluation datasets. The datasets are organized to test models on a spectrum of complexity, from context-free single-turn queries to intricate, real-world multi-turn dialogues.

The prefixes \texttt{ST-} and \texttt{MT-} in the table headers distinguish between these evaluation contexts:
\begin{itemize}
    \item \textbf{ST (Single-Turn):} Refers to evaluations on isolated queries where no conversation history is provided. This context is used for the CLINC150 dataset (Table \ref{clinc_table}).
    \item \textbf{MT (Multi-Turn):} Refers to evaluations on queries within a dialogue, where understanding prior context is essential for correct routing. This context applies to the MANtIS, SGD, and LMSYS datasets (Tables \ref{mantis_label}, \ref{sgd_table}, and \ref{lmsys_table}).
\end{itemize}

\begin{table}[ht]
\centering
\begin{tabular}{lccccc}
\toprule
\textbf{Dataset} & \textbf{\# Domains} & \textbf{\# Actions} & \textbf{\# Conversations} & \textbf{\# Spans} & \textbf{\# Turns} \\
\midrule
Training   & 981 & 3\,156 & 5\,194 & \textit{N/A} & 43\,442 \\
CLINC150   &   8 &   80  & \textit{N/A} & \textit{N/A} & 1\,000 \\
MANtIS     &  14 & \textit{N/A} & \textit{N/A} & \textit{N/A} & 1\,000 \\
SGD        &  12 &  252 & 2\,500 & 2\,000 & 5\,141 \\
LMSYS-1M   & 258 &  390 &   123 &   662 &   821 \\
\bottomrule
\\
\end{tabular}

\caption{Dataset statistics}
\label{table:dataset_statistic}
\end{table}
\begin{table}[ht]
\centering
\begin{tabular}{lccc}
\toprule
\textbf{Model} & \textbf{ST-fQfA} & \textbf{ST-fQcA} & \textbf{Irrelevance} \\
 & Turn & Turn & Turn \\ 

\midrule
Qwen2.5-1.5B                    & 67.80 & 50.30 &  9.40 \\
\midrule
GPT-4o                       & \textbf{97.20} & 94.80 & \textbf{96.20} \\
GPT-4o-mini                  & 94.80 & 93.50 & 95.40 \\
Claude-Sonnet-3.7            & 95.10 &\textbf{96.20} & 95.60 \\
Claude-Haiku-3.5             & 91.00 & 95.80 & 87.20 \\
Gemini 2.0 Flash             & 91.00 & 95.80 & 87.20 \\
Gemini 2.0 Flash-Lite        & 94.80 & 94.10 & 85.00 \\
\midrule
\textbf{Arch-Router}                  & \textbf{97.20} & 94.10 & 95.20 \\
\bottomrule
\\
\end{tabular}
\caption{Single-turn routing accuracy - CLINC150}
\label{clinc_table}
\end{table}

\begin{table}[ht]
\centering

\begin{tabular}{lc}
\toprule
Models & \textbf{MT-fQcA} \\ 
 & Turn \\ \midrule
 Qwen2.5-1.5B                & 40.40 \\
\midrule
GPT-4o                   & 88.40 \\
GPT-4o-mini              & 83.50 \\
Claude-Sonnet-3.7        & \textbf{91.30} \\
Claude-Haiku-3.5         & 87.50 \\
Gemini 2.0 Flash         & 83.90 \\
Gemini 2.0 Flash-Lite    & 82.20 \\
\midrule
\textbf{Arch-Router}                & 90.10 \\

\bottomrule \\
\end{tabular}
\caption{Multi-turn, single route in conversation history - MANtIS}
\label{mantis_label}

\end{table}


\begin{table}[ht]
\centering
\begin{tabular}{lccccccccc}
\toprule
\multirow{2}{*}{\textbf{Models}} &
\multicolumn{3}{c}{\textbf{MT-fQfA}} &
\multicolumn{3}{c}{\textbf{MT-fQcA}} &
\multicolumn{2}{c}{\textbf{Irrelevance}} \\
\cmidrule(lr){2-4}\cmidrule(lr){5-7}\cmidrule(lr){8-9}
 & Turn & Span & Conv. & Turn & Span & Conv. & Turn & Span \\
\midrule
Qwen2.5-1.5B                 & 63.07 & 34.95 & 34.20 & 30.67 & 22.68 &  6.63 & 12.90 &  8.40 \\
\midrule
GPT-4o                    & \textbf{99.81} & 92.90 & 92.25 & 95.63 & 88.66 & 87.29 & 83.54 & 84.50 \\
GPT-4o-mini               & \textbf{99.81} & 92.20 & 92.25 & 83.09 & 80.15 & 75.70 & 51.33 & 46.94 \\
Claude-Sonnet-3.7         & 99.41 & \textbf{97.40} & 93.40 & 95.63 & 89.04 & 87.29 & \textbf{97.40} & \textbf{96.40} \\
Claude-Haiku-3.5          & 98.59 & 95.40 & 91.25 & 92.52 & 87.80 & 82.87 & 63.00 & 62.10 \\
Gemini 2.0 Flash          & 97.53 & 91.20 & 91.00 & 90.96 & 83.44 & 80.56 & 76.80 & 76.50 \\
Gemini 2.0 Flash-Lite     & 93.60 & 86.30 & 88.40 & 89.92 & 77.88 & 68.51 & 67.90 & 62.30 \\
\midrule
\textbf{Arch-Router}                 & 99.20 & 97.00 & \textbf{94.60} &\textbf{96.26} &\textbf{91.68} & \textbf{90.05} & 95.40 & 95.20 \\

\bottomrule \\
\end{tabular}
\caption{Multi-turn, multi-route in history - SGD}
\label{sgd_table}

\end{table}

\begin{table}[ht]
\centering
\begin{tabular}{lcccccccc}
\toprule
\multirow{2}{*}{\textbf{Models}} &
\multicolumn{3}{c}{\textbf{MT-fQfA}} &
\multicolumn{3}{c}{\textbf{MT-fQcA}} &
\multicolumn{2}{c}{\textbf{Irrelevance}} \\
\cmidrule(lr){2-4}\cmidrule(lr){5-7}\cmidrule(lr){8-9}
 & Turn & Span & Conv. & Turn & Span & Conv. & Turn & Span \\
\midrule
Qwen2.5-1.5B                  & 24.36 & 18.12 &  4.80 & 12.18 &  6.04 &  0.80 &  9.13 &  6.37 \\
\midrule
GPT-4o                     & 94.28 & 93.25 & 81.30 & 88.58 & 87.13 & 77.24 & \textbf{99.08} & \textbf{98.73} \\
GPT-4o-mini                & 92.69 & 90.75 & 78.86 & 84.04 & 80.97 & 73.17 & 95.89 & 93.63 \\
Claude-Sonnet-3.7          & 96.10 & 95.27 & 84.55 & \textbf{94.76} & \textbf{91.99} & \textbf{84.55} & 98.63 & 98.09 \\
Claude-Haiku-3.5           & 91.96 & 88.97 & 80.49 & 85.26 & 80.82 & 77.24 & 91.32 & 87.26 \\
Gemini 2.0 Flash           & 92.69 & 91.33 & 80.49 & 84.23 & 83.13 & 75.61 & 91.32 & 89.17 \\
Gemini 2.0 Flash-Lite      & 90.38 & 90.08 & 65.87 & 73.08 & 61.93 & 57.94 & 71.69 & 70.06 \\
\midrule
 \textbf{Arch-Router}             & \textbf{97.93} & \textbf{96.83} & \textbf{86.99} & 93.79 & 90.63 & 82.29 & 98.87 & 98.54 \\
\bottomrule
\\
\end{tabular}

\caption{Multi-turn, multi-route with diverse conversation - LMSYS}
\label{lmsys_table}

\end{table}

\section{Latency Analysis}

\begin{table}[ht]
    \centering
    
    \begin{tabular}{lccc}
        \toprule
        \textbf{Model}  & \textbf{Latency (ms)} & \textbf{Performance (\%)} \\
        \midrule
        
        GPT-4o                          & $836 \pm 239$  & 89.74          \\
        GPT-4o-mini                     & $737 \pm 164$  & 82.79          \\
        Claude-sonnet-3.7              & $1450 \pm 385$ & 92.79 \\
        Claude-haiku-3.5               & $1249 \pm 352$ & 84.96          \\
        Gemini-2.0-flash                & $581 \pm 101$  & 85.63          \\
        Gemini-2.0-flash-lite         & $510 \pm 82$   & 76.69          \\
        \midrule
        \textbf{Arch-Router}  & $\mathbf{51 \pm 12}$   & \textbf{93.17}         \\
        \bottomrule
        \\
    \end{tabular}
    \caption{Performance and Latency Analysis of Router Models. The table compares latency (average ± standard deviation in milliseconds) benchmarked from OpenRouter ), and overall routing performance. The cost for Arch-Router is estimated in hosting the model on AWS L40S instance.}
    \label{tab:cost_performance_analysis}
\end{table}

The result clearly shows Arch-Router's main advantage: its incredible efficiency with near instance latency. Arch-Router achieves better accuracy as the best commercial model, Claude-sonnet-3.7 (93.17\% vs. 92.79\%) while also has a massive gap in latency. Arch-Router is over  28 times faster than its closest competitor. Arch-Router offers a unique solution by delivering the accuracy of a large, premium model with the latency of a small, efficient one, making it ideally suited for industry applications.

\paragraph{Latency Benchmark.}
We measured the end-to-end completion latency, representing the total time from request to full response for each model. The benchmark was performed on 2,000 routing prompts randomly selected from our training dataset. Commercial model latencies were measured via the OpenRouter API, while Arch-Router was benchmarked on its self-hosted AWS L40S instance to reflect its practical deployment speed.

\paragraph{Disclaimer.} It is important to note that the latency presented for commercial models are based on snapshots from the OpenRouter API at the time of our evaluation and are subject to change. The performance of these third-party services can vary based on server load, network conditions, and evolving pricing structures. However, while the exact figures may fluctuate, the fundamental conclusion of our analysis remains robust. The orders-of-magnitude difference in speed between a compact 1.5B model like Arch-Router and large, general-purpose APIs is a structural advantage. 
\section{Case Study: Routing in a Coding Session}
\label{sec:case-study}

\subsection{Conversation and Routes}
A multi-turn coding interaction (11 user turns) was routed by \textbf{Arch-Router} a using the prompt template in Table \ref{tab:system-prompts} and RouteLLM.  Table~\ref{tab:coding_routing} lists each \emph{user request}, the \emph{route policy} chosen by the router, and whether that choice matched the ground-truth intent. The route policies is shown in Table \ref{tab:coding-routes} which provide policy name, description, and corresponding LLM. The model choice is purely based on each our preferences.

\begin{table}[ht]
\centering\small
\begin{tabular}{@{}L{3.6cm}L{7cm}L{3.2cm}@{}}
\toprule
\textbf{Route Policy} & \textbf{Description} & \textbf{ LLM}\\
\midrule
\texttt{code\_generation} & Generate new code snippets, functions, or boiler-plate from user requirements. & Claude-sonnet-3.7\\
\texttt{code\_explanation} & Explain what a piece of code does, including logic and edge cases. & GPT-4o\\
\texttt{bug\_fixing} & Identify and fix errors or bugs in user-supplied code. & GPT-4o\\
\texttt{performance\_optimization} & Suggest changes to make code faster, more scalable, or more readable. & GPT-4o\\
\texttt{api\_help} & Assist with understanding or integrating external APIs and libraries. & GPT-4o-mini\\
\texttt{programming\_question} & Answer general programming theory or best-practice questions. & GPT-4o-mini\\
\texttt{default} & & Qwen2.5-4B \\
\bottomrule
\\
\end{tabular}
\caption{Route policy for the coding application.}
\label{tab:coding-routes}
\end{table}

\subsection{Analysis}
Across the conversation Arch-Router perfectly routed \textbf{8 / 8} user turns. Its generative architecture digests the \emph{entire} conversation plus the natural language descriptions, so it can interpret elliptical follow-ups. After producing code in Turn 2, it recognized “\emph{this doesn’t work, try again}” as a \texttt{bug\_fixing} request—even though no keywords such as “bug” or “fix” appeared—and later mapped “\emph{run too slow}” to \texttt{performance\_optimization}. Because new routes are just additional text in the routing policy, the same mechanism works if a product team inserts, say, \texttt{security\_hardening} or \texttt{accessibility\_audit} without retraining.

RouteLLM’s policy is trained to predict answer \emph{quality scores} from benchmarks, then pick a “weak” or “strong” model accordingly. That shortcut works when each query is self-contained, but it breaks as soon as meaning depends on prior turns. In our conversation, RouteLLM downgraded three context-dependent requests (Turns 3, 5, 7) to a cheap model, weaker model. The mis-routes illustrate a structural limitation:  it is uncertain whether a follow-up will yield a costly code generation, a quick bug fix, or simply a thank you response.

\begin{table}[ht]
\centering\small
\begin{tabular}{@{}c p{5.5cm} p{2cm} p{4cm}}
\toprule
\textbf{Turn} & \textbf{User request} & \textbf{RouteLLM} & \textbf{Arch-Router}  \\
\midrule
1 & Hi & \texttt{weak} \cmark & \texttt{general} \cmark   \\
2 & Write a function to visualize an dataframe that has error column, the visualization the accuracy aggregation over all the rows & \texttt{strong} \cmark& \texttt{code\_generation} \cmark \\
3 & This doesnt work& \texttt{weak} \xmark & \texttt{bug\_fixing}\cmark   \\
4 & Okay, now try the same thing for a dataframe with a label, output column, please check the correctness of the output column compared to the label, note that there is different t incorrect types: relevance, irrelevance, and incorrect format & \texttt{strong} \cmark& \texttt{code\_generation}\cmark   \\
5 & The code runs too slow, any way u can simplify it to make it faster? & \texttt{weak}  \xmark& \texttt{performance\_optimization} \cmark  \\
6 & What are the functions that can be replaced from seaborn & \texttt{strong} \cmark & \texttt{api\_help} \cmark  \\
7 & Any other ones? & \texttt{weak} \xmark& \texttt{api\_help} \cmark \\
8 & Thats all, thank you & \texttt{weak} \cmark & \texttt{general} \cmark  \\
\bottomrule
\\
\end{tabular}
\caption{Comparison of routing predictions from RouteLLM and Arch-Router for each user turn in a coding conversation.}
\label{tab:coding_routing}
\end{table}

\end{document}